\documentclass[10pt,conference]{IEEEtran}
\usepackage{amsmath,amssymb,amsfonts}
\usepackage{algorithmic}
\usepackage{graphicx}
\usepackage{textcomp}
\usepackage{xcolor}
\usepackage{array}
\usepackage{multirow}
\usepackage{subcaption}
\usepackage[
backend=biber,
sorting=nty
]{biblatex}

\begin{document}

\title{Attempt Towards Stress Transfer in Speech-to-Speech Machine Translation}

\author{
\textit{Sai Akarsh, Vamshi Raghusimha, Anindita Mondal, Anil Vuppala}\\
LTRC, International Institute of Information Technology - Hyderabad\\
\texttt{{\footnotesize\{sai.akarsh, narasinga.vamshi, anindita.mondal\}@research.iiit.ac.in, anil.vuppala@iiit.ac.in}}}

\maketitle

\begin{abstract}
The language diversity in India's education sector poses a significant challenge, hindering inclusivity. Despite the democratization of knowledge through online educational content, the dominance of English, as the internet's lingua franca, limits accessibility, emphasizing the crucial need for translation into Indian languages. Despite existing Speech-to-Speech Machine Translation (SSMT) technologies, the lack of intonation in these systems gives monotonous translations, leading to a loss of audience interest and disengagement from the content. To address this, our paper introduces a dataset with stress annotations in Indian English and also a Text-to-Speech (TTS) architecture capable of incorporating stress into synthesized speech. This dataset is used for training a stress detection model, which is then used in the SSMT system for detecting stress in the source speech and transferring it into the target language speech. The TTS architecture is based on FastPitch and can modify the variances based on stressed words given. We present an Indian English-to-Hindi SSMT system that can transfer stress and aim to enhance the overall quality and engagement of educational content.
\end{abstract}
\begin{IEEEkeywords} speech-to-speech machine translation, stress detection, text-to-speech, speech synthesis\end{IEEEkeywords}

\section{Introduction}
SSMT stands at the forefront of linguistic innovation, aiming to bridge communication gaps by offering real-time translation of spoken language. In a globalized society, effective communication across diverse linguistic boundaries is indispensable. SSMT, as a cutting-edge technology, emerges as a pivotal solution and is composed of three primary components: Automatic Speech Recognition (ASR), Machine Translation (MT), and Text-to-Speech synthesis (TTS). These components involve language and prosodic understanding and hence, there are several challenges, stemming from the complex nature of spoken language and the intricacies involved in real-time translation.\par
Prosody encompasses elements such as stress, rhythm, and intonation, which play a crucial role in conveying the emotional and expressive aspects of spoken language. Ensuring the faithful transfer of prosody from the source to the target language is essential for producing natural and human-like translations. Prosody often varies based on the context of the conversation, such as asking a question, making a statement, or expressing surprise. The inability to transfer prosodic features accurately results in robotic or unnatural-sounding translations, affecting the overall intelligibility and expressiveness of the translated speech.\par
Traditional speech translation systems overlook paralinguistic cues, but there has been a recent focus on systems that transfer linguistic content along with emphasis information. In contrast to conventional SSMT systems \cite{emphasis-s2st, emphasis-seq2seq, emphasis-attention}, proposed methods to translate paralinguistic information, focusing specifically on stress/emphasis where source speech was English and target speech was Japanese. Utilizing DNN-HMM ASR, Statistical Machine Translation, and an HSMM-based TTS model, the approach in \cite{emphasis-s2st} involves the use of linear-regression hidden semi-Markov models (LRHSMMs). These models estimate a sequence of real-numbered emphasis values for each word in an utterance.\par
Numerous studies have been conducted within the linguistic domains of Spanish, English, and Catalan such as \cite{pros-s2st},\cite{holistic-s2st} reflecting a comprehensive exploration of prosody transfer techniques between these languages. Moreover, recent research initiatives have broadened their focus to incorporate the intricate linguistic environment prevalent in the Indian context \cite{vakta-setu}. While existing speech datasets offer valuable resources for speech processing and synthesis research, they often lack annotations for aspects like prosodic emphasis, particularly in domains such as educational settings. To address this gap, this paper undertook an in-depth exploration of an Indian English dataset \cite{ied-iiith}, meticulously annotated the gathered data, devised a stress detection model, and subsequently formulated a methodology to integrate this model into TTS systems which can generate target language speech in Hindi enriched with stress patterns for enhanced naturalness and expressiveness.

\section{Proposed Methodology}
The fidelity of SSMT hinges not just on semantic accuracy, but also on capturing the speaker's intent conveyed through subtle prosodic cues like stress. While existing systems excel at translating meaning, they often stumble when replicating the emotional weight and emphasis embedded in the source speech. In this section, we discuss in detail the methodology for bridging this gap.

\subsection{Stress Dataset}
Recognizing the limitations of existing speech datasets in capturing the subtleties of stress, especially within the distinct domain of Indian English video lectures, we sought to build a resource tailored to this specific niche. We leveraged the existing strengths of the IED-IIITH dataset \cite{ied-iiith}, which features high-quality speech recordings from the NPTEL educational lectures. This dataset was chosen as it contains a wide variety of speakers and speaking styles, and also captures common disfluencies that occur in spoken speech. This is important as our stress detection model which is set in the context of spoken speech in educational lectures needs to be robust against disfluencies and shouldn't necessarily classify them as stress regions.\par

In the context of lectures, the lecturer naturally incorporates stress while explaining specific topics, emphasizing certain words to add meaning and subtlety. Such stress patterns often occur at the beginning or ending of sentences, and feature prolonged duration, with potential pauses. This dataset was manually annotated by trained annotators who identified the stressed regions, relying on both acoustic cues like pitch and duration and indicators like waveform and the above-explained reasoning.\par

\begin{figure}[h]
    \centering
    \includegraphics[width=1\linewidth]{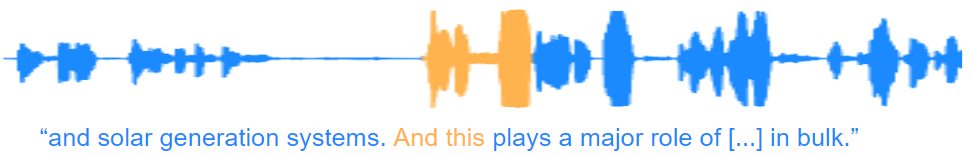}
    \caption{Sample from the annotated dataset that contains a stress region (orange)}
    \label{figure:stress_marking}
\end{figure}

Label Studio \cite{label_studio}, an open-source data labelling platform, was used to perform stress region annotation and generate the timestamps for each audio file as shown in Figure \ref{figure:stress_marking} (with the transcript for reference). An audio file may have multiple stressed regions or none at all. The audio data was distributed among ten annotators in such a way that every audio sample had labels from at least 3 different annotators. This was done to address the subjective nature of stress and emphasis as everyone has a different perception of such nuances. We use Fleiss Kappa inter-annotator agreement \cite{fleiss-kappa} to aggregate multiple annotations to get the final labels. This effort resulted in a unique stress-annotated dataset specifically tailored to Indian English (NPTEL) video lectures, advancing the availability of resources for stress detection.

\subsection{Stress Detection Models}
Stress detection in speech is achieved by training classifiers on several acoustic features calculated from the data collected. The classifier is trained to predict a stress label for each frame indicating if a frame is stressed or not. The model uses features from a window of frames rather than one frame for more contextual information. In speech, the stressed regions occupy a small region when compared to the non-stressed regions, hence the labels given to the classifier are usually skewed giving rise to more false positives and need to be balanced.\par
A post-processing technique is employed, where the word alignments with speech are calculated using an ASR system. With these alignments, the frame-level predictions are converted to word-level predictions. A word is classified as stressed if the region containing the word has a majority of frames predicted as stress. This technique not only filters out a considerable amount of false positives but also gives stress detection on the word level which is used in SSMT.\par
For every detected stress word, additional information, indicating how the pitch and energy variances change in that stressed region with respect to the rest of the speech, is denoted as a scaling factor and is quite useful in SSMT as a metric for the TTS system to modify variances in order to introduce stress in the target speech. 

\subsection{TTS}
Once the stressed words are known, a way of incorporating stress in synthesized speech is presented based on FastPitch \cite{łańcucki2021fastpitch} which is a two-stage TTS architecture. The first stage is comprised of a Transformer-based \cite{attention-transformers} text encoder, and a CNN-based pitch and duration variance predictor. The second stage combines the text encoding with pitch embeddings and duplicates it based on the predicted duration, which is then sent to another Transformer to finally give the mel spectrogram.\par

\begin{figure*}[t!]
    \centering
    \begin{subfigure}[t]{0.7\textwidth}
        \centering
        \includegraphics[height=2.7in]{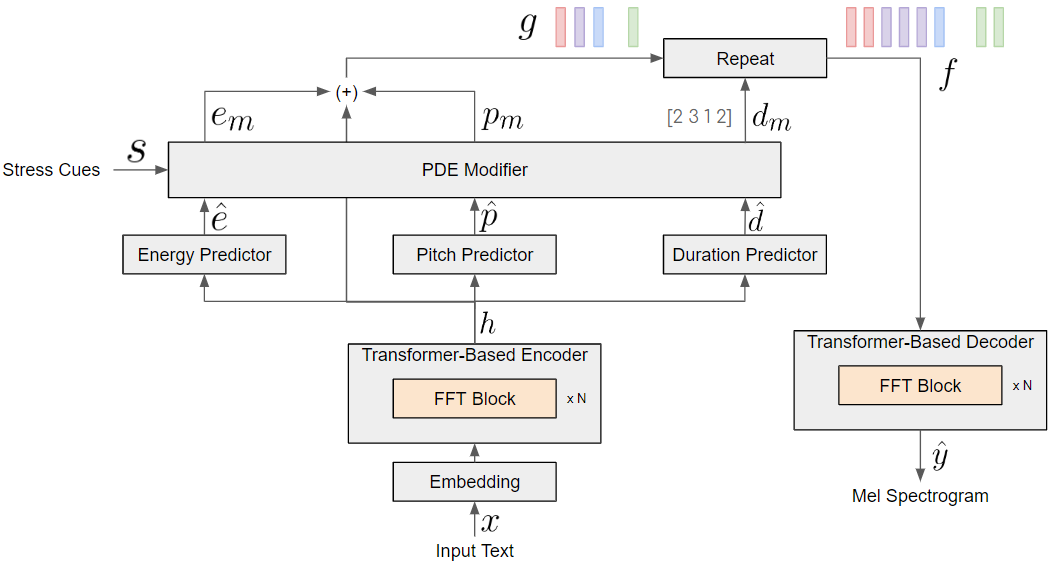}
        \caption{Proposed TTS architecture}
    \end{subfigure}%
    ~ 
    \begin{subfigure}[t]{0.3\textwidth}
        \centering
        \includegraphics[height=1.9in]{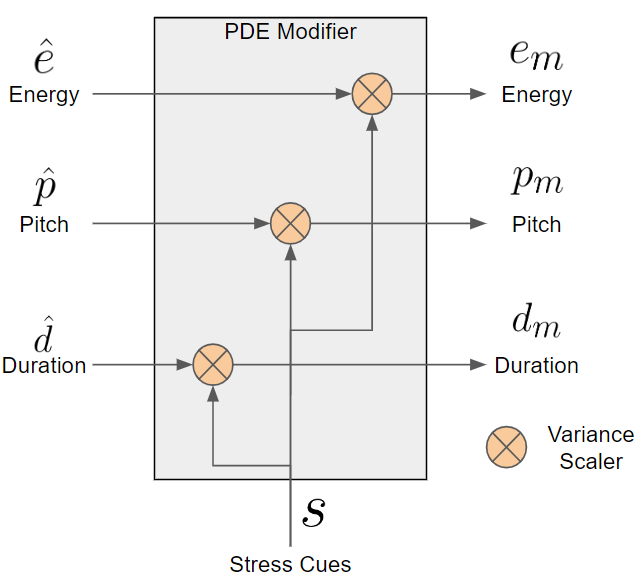}
        \caption{PDE Modifier}
    \end{subfigure}
    \caption{\textbf{Architecture for proposed TTS} follows FastPitch \cite{łańcucki2021fastpitch}. The predicted variances are modified by the PDE Modifier based on the stress cues to introduce stress in the synthesized speech.}
    \label{figure:tts_arch}
\end{figure*}

In the proposed TTS system, several changes were made to the base FastPitch architecture to condition the synthesized speech on stress cues. First, both text and stress cues are taken as input, where the text is handled as character tokens and stress cues are given as stressed word and scale factor pairs that were generated using the stress detection model. Second, an energy prediction block is added since stressed words have a strong correlation with energy and can be seen in Figure  \ref{figure:tts_arch} (a). Third, a Pitch-Duration-Energy (PDE) Modifier block as shown in Figure \ref{figure:tts_arch} (b) is introduced, which modifies the predicted variances using the stress cues. Such modifications enable the training of the TTS on neutral speech-text datasets which are already available. Since TTS requires very high-quality clean speech data, procuring such amounts of stressed data is not trivial and we overcome this issue by using the PDE Modifier block only during inference.\par

Let $x$ be input character tokens and $s$ be input stress cues. The token embeddings for the text are generated using an embedding layer and are passed to a feed-forward Transformer (FFT) encoder stack that produces the hidden representation $h=\text{FFT}_e\text{(Embedding}(x))$. This hidden representation is then used to predict each input character token's duration $\hat{d}$, pitch $\hat{p}$ and energy $\hat{e}$ using a CNN stack. The pitch, duration and energy values are then sent to the PDE Modifier block, where each of the variances is modified based on the stressed word region and the scaling factors provided. This is done by scaling the predicted variances of every input token corresponding to a stressed word using the scaling factors.\par
Next, the modified pitch $p_m$ and energy $e_m$ are projected to match the dimensionality of the hidden representation $h$ using an embedding layer and are then added to $h$. The resulting sum $g$ which is at the resolution of input character tokens is then discretely upsampled using the modified durations $d_m$ to give $f$ which is at the resolution of output frames. $f$ is then passed to an FFT decoder stack to produce the output mel-spectrogram sequence.\par
The output mel-spectrogram $\hat{y}$ is generated using the modified variances from the PDE Modifier block, thus incorporating stress in the synthesized speech. Since the PDE Modifier block is used only during inference, the training loss is based on mean-square error (MSE), similar to that of FastPitch.

\begin{figure*}[t!]
    \centering
    \includegraphics[width=1\linewidth]{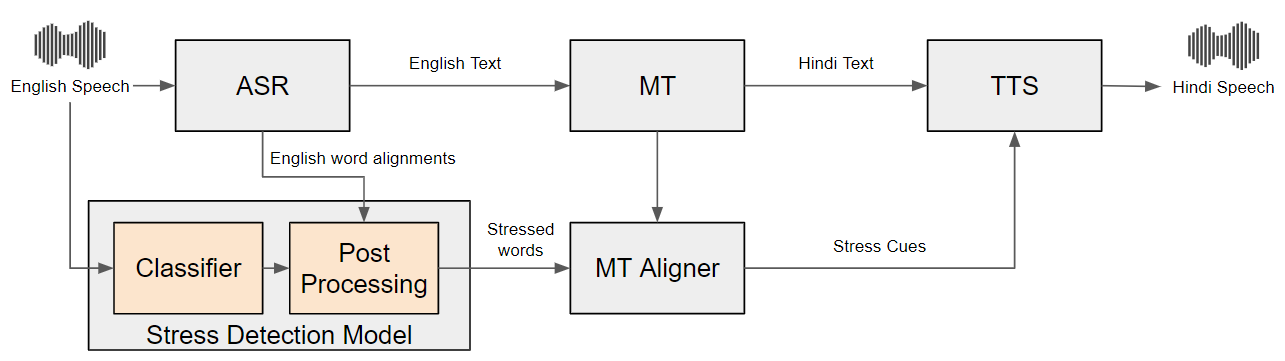}
    \caption{\textbf{The overall architecture for the proposed SSMT pipeline}. The MT Aligner takes English stress words, generates the English-Hindi word alignments and converts them into stress cues: Hindi stress words along with scaling factors}
    \label{figure:ssmt_arch}
\end{figure*}

\section{Experimental Setup}
In this section, we discuss the datasets, features and models that were used for the experiments.

\subsection{Stress Dataset}
The stress dataset consists of nearly 10 hours of spoken speech in the context of educational video lectures from \cite{ied-iiith}. It features recordings from 60 speakers (30 male and 30 female), where each speaker has around 10 minutes of speech. These 10-minute recordings are further split into smaller audio files of 8 to 12 seconds, with a sampling rate of 16000 Hz. The dataset comprises 3329 audio files where each audio file is identified using a unique ID and has information about the speaker and the stressed region labels. Each audio sample can have multiple stressed regions or none at all. There are a total of 3807 occurrences of stressed regions with an average duration of about 0.53 seconds.

\begin{table*}[h]
    \centering
    \begin{tabular}{|c|c|c|c|c|c|c|c|c|c|c|}
        \hline
        \multirow[t]{3}{*}{\textbf{Features}} & \multirow[t]{3}{*}{\textbf{Models}} & \multicolumn{3}{|c|}{\textbf{Window Size = 3}} & \multicolumn{3}{|c|}{\textbf{Window Size = 5}} & \multicolumn{3}{|c|}{\textbf{Window Size = 7}} \\
        \cline{3-11}
        & & \textbf{Accuracy} & \textbf{F1} & \textbf{Post Accuracy} & \textbf{Accuracy} & \textbf{F1} & \textbf{Post Accuracy} & \textbf{Accuracy} & \textbf{F1} & \textbf{Post Accuracy} \\
        \hline
        & & & & & & & & & & \\
        F0 and Energy & LPA & 60.6 & 60.45 & 71.11 & \textbf{68.9} & \textbf{68.77} & \textbf{81.85} & \textbf{75.3} & \textbf{75.16} & \textbf{86.18} \\
        & RFC & 58.2 & 58.06 & 69.52 & 68.6 & 67.59 & 76.75 & 72.8 & 71.79 & 82\\
        & SVC & \textbf{61.1} & \textbf{60.5} & \textbf{73.35} & 67.2 & 66.11 & 80.3 & 71.5 & 70.46 & 83.83 \\
        \hline
        & & & & & & & & & &\\
        F0, Energy, MFCC, SDC & LPA & 63.1 & 61.09 & 75.28 & 62.2 & 58.52 & 74.57 & \textbf{82.3} & \textbf{80.77} & \textbf{90.36} \\
        & RFC & 64.3 & 63.28 & 77.37 & \textbf{71.1} & \textbf{70.3} & \textbf{81.22} & 78.3 & 78.27 & 85.29 \\
        & SVC & \textbf{66.8} & \textbf{66.79} & \textbf{78.42} & 70.1 & 70.09 & 80.12 & 79.2 & 79.2 & 83.82 \\
        \hline
    \end{tabular}
    \caption{Evaluation of stress detection model with different features and varying window sizes.}
    \label{tab:features_models}
\end{table*}

\subsection{Feature Extraction and Stress Detection Models}
Several acoustic features such as F0 \cite{Fujisaki}, energy \cite{num2num}, Mel Frequency Cepstral Coefficients (MFCC) and Shifted Delta Coefficients (SDC) \cite{ied-iiith} were used for the task of stress detection. 13 MFCC features were extracted by taking a frame length of 1024 and hop length of 256. 52 SDC features were calculated using $d=1, p=5, k=3$ as they gave better results. The F0 and energy contours were averaged to get frame-level features. All the features were then mean-variance normalized, and different subsets of features were considered to get the best performance. We combine different features on the frame level and then stack up those features from frames before and after to give wider contextual information.\par
The data imbalance needs to be tackled as the frames corresponding to stressed regions are much fewer than that of non-stressed regions. This would lead the model to be biased and hence not desirable. We used SMOTE (Synthetic Minority Oversampling Technique) \cite{smote_chawla} based oversampling on the data while training each model.\par
Using the features extracted above, various classifiers were trained to perform the task of stress detection. The following three classifiers were chosen to be tabulated as they gave better results: Label Propagation Algorithm (LPA) \cite{label_prop} with radial basis function kernel and 7 neighbours, Random Forest Classifier (RFC) with 100 trees in the ensemble and Support Vector Classifier (SVC) with a radial basis function kernel and penalty factor of 0.8. Due to the limited availability of data, a DNN-based model was not deemed suitable to achieve the desired level of accuracy.

\subsection{SSMT Pipeline}
\textbf{ASR:} WhisperX \cite{bain2022whisperx} ASR model was used to process the source speech in Indian English to give both the transcript and word alignments. The ASR is built upon a pre-trained Whisper \cite{whisper} English large-v2 model and delivers accurate word-level transcriptions with high confidence scores. Word-level alignments in post-processing boost stress detection model accuracy and reliability. They also aid in identifying stressed words in the target language by mapping them to their equivalents in the source language. \par
\textbf{MT and Aligner:} English-Hindi MT model was taken from the open source Helsinki OPUS MT\cite{HelsinkiNLP-OPUSMT} project. This is used to convert the English transcript from ASR into a Hindi transcript. MT word aligner from simalign \cite{simalign} was used to map the translated Hindi words to their corresponding English words. It uses a pre-trained mBERT \cite{devlin2019bert} multilingual language model to achieve this mapping. \par
\textbf{TTS:} To train the TTS model, we have used the latest version of the Hindi Male speaker's database from the ULCA Bhashini IndicTTS \cite{indictts-baby} which has nearly 10 hours of recorded and transcribed speech at a sampling rate of 22050 Hz. The utterances of more than 15 seconds were ignored. The TTS model used an FFT stack of 6 layers and a pre-trained Tacotron2 \cite{tacotron2} model for calculating the durations of ground truth \cite{łańcucki2021fastpitch}. It was trained for 2000 epochs on a single NVIDIA A100 80GB Tensor Core GPU with a batch size of 128.

\section{Results}
The performance of the stress detection models is shown in Table \ref{tab:features_models} based on the classification of frame-level stress labels. The experiments were carried out on two feature sets: (F0, Energy) and (F0, Energy, MFCC, SDC) using different classifiers. From Table \ref{tab:features_models}, it can be seen that LPA having a window size of 7 with the addition of the post-processing stage improves the accuracy and f1 score for both feature sets with an absolute improvement of around 2-4\%. RFC gave better results compared to LPA and SVC in the case of all features when the window size is 5 with an absolute improvement of around 5\%. SVC gave better results compared to LPA and RFC in both cases when the window size is 3 with an absolute improvement of around 2-3\%.\par

\begin{table}[h]
    \centering
    \begin{tabular}{|c|cc|}
        \hline
        \multicolumn{1}{|l|}{} & \multicolumn{2}{c|}{\textbf{MOS}}                               \\ \hline
        \textbf{TTS Model}     & \multicolumn{1}{c|}{\textbf{Without Stress}} & \textbf{With Stress} \\ \hline
        Pitch                  & \multicolumn{1}{c|}{\textbf{4.25}}             & 3.95                     \\ \hline
        Pitch and Energy       & \multicolumn{1}{c|}{\textbf{4.36}}             & 4.21                     \\ \hline
    \end{tabular}
    \caption{Evaluation of TTS models in SSMT using Mean Opinion Score (MOS)}
    \label{tab:tts_mos}
\end{table}

Due to the lack of automated evaluation metrics for SSMT, we rely on subjective evaluation through a survey where 30 participants, fluent in both English and Hindi were asked to give performance metrics. Every participant was given 15 samples\footnote{https://sativus04.github.io/}, each containing the original English speech, translated Hindi speech with and without stress. They gave two performance metrics, the first is the quality of the synthesized Hindi speech for both with and without stress as seen in Table \ref{tab:tts_mos}. A score between 0 and 5 is given, where 0 is very bad and 5 is very good. The Mean Opinion Score (MOS) is derived by averaging all the scores. Notably, TTS without stress outperforms its stressed counterpart due to potential instability in spectrogram generation when directly altering variances. The model with both Pitch and Energy exhibits enhanced flexibility in speech generation.\par

\begin{table}[h]
    \centering\begin{tabular}{|c|c|}
        \hline
        \textbf{SSMT}    & \textbf{st-MOS} \\ \hline
        Pitch            & 3.73           \\ \hline
        Pitch and Energy & \textbf{4.09}           \\ \hline
    \end{tabular}
    \caption{Evaluation of SSMT models using st-MOS (Stress Transfer MOS)}
    \label{tab:ssmt_mos}
\end{table}

The second performance metric is the measure of how well the stress has been incorporated as seen in Table \ref{tab:ssmt_mos}. The participant uses the original speech and synthesized speech without stress as a reference to gauge how well the stress has been added to the synthesized speech with stress. This is used to calculate the st-MOS (stress transfer MOS) score, where 0 means that no stress was transferred and 5 means that all the stress regions were transferred. The average score of 3.96 indicates effective stress integration, particularly with both pitch and energy, underlining the significance of energy in stress modelling. These results serve as the baseline for this task, given the absence of a baseline setup. \par


\section{Conclusion}
SSMT systems keep improving with rapid strides. We curated a dataset with stress annotations for Indian English and trained stress detection models using it. We modified an existing TTS architecture for the addition of stress in synthesized speech using existing speech corpora. Evaluation was done on the stress detection models, TTS and SSMT systems and the best configurations were identified. Several directions of work can be seen from here: other TTS models need to be compared in this task, preserving the speaker's voice in target speech, and designing better subjective and objective metrics for such tasks in SSMT. We aim to address some of these issues in future works and bridge the gap between the source and target speech in SSMT.
\section*{Acknowledgment}
\addcontentsline{toc}{section}{Acknowledgment}
We thank the reviewers for their insightful comments. This undertaking is funded by the Ministry of
Electronics and Information Technology, Government of India
, as evidenced by the Sanction Order:
11(1)/2022-HCC(TDIL)-Part(2)/A/B/C and the Administrative Approval: 11(1)/2022-HCC(TDIL)-
Part(2).

\printbibliography
\end{document}